\documentclass[10pt,twocolumn,letterpaper]{article}

\usepackage{cvpr}
\usepackage{times}
\usepackage{epsfig}
\usepackage{graphicx}
\usepackage{amsmath}
\usepackage{amssymb}
\usepackage{epigraph}
\usepackage{authblk}

\usepackage{graphicx}
\usepackage{mathtools}
\usepackage{multirow}
\usepackage{commath}
\usepackage{caption}
\usepackage{xcolor}
\usepackage{authblk}
 
\definecolor{purple}{rgb}{0.3, 0, 0.51}

\usepackage{hyperref}

\cvprfinalcopy % *** Uncomment this line for the final submission

\ifcvprfinal\fi
\begin{document}
%%%%%%%%% TITLE
\title{PREDICT \& CLUSTER: Unsupervised Skeleton Based Action Recognition}
\author[1]{Kun Su}
\author[1]{Xiulong Liu}
\author[1, 2]{Eli Shlizerman}

\affil[1]{Department of Electrical \& Computer Engineering, University of Washington, Seattle, USA}
\affil[2]{Department of Applied Mathematics, University of Washington, Seattle, USA}

% For a paper whose authors are all at the same institution,
% omit the following lines up until the closing ``}''.
% Additional authors and addresses can be added with ``\and'',
% just like the second author.
% To save space, use either the email address or home page, not both
% \and
% Second Author\\
% Institution2\\
% First line of institution2 address\\
% {\tt\small secondauthor@i2.org}

% \teaser{
%     	\includegraphics[width=.85\textwidth{Fig0.pdf} 
%     	\caption{ PREDICT \& CLUSTER: Unsupervised action recognition from body keypoints.} 
%     	\label{fig:teaser}
% }

% \makeatletter
% \let\@oldmaketitle\@maketitle% Store \@maketitle
% \renewcommand{\@maketitle}{\@oldmaketitle% Update \@maketitle to insert...
% \centering
% \includegraphics[width=0.95\linewidth]{Fig0.pdf}\\
% \caption{PREDICT \& CLUSTER: Unsupervised action recognition from body keypoints.}
% \label{fig:teaser}
% }% ... an image
% \makeatother
% \maketitle

\twocolumn[{
\renewcommand\twocolumn[1][]{#1}
\maketitle
\begin{center}
    \includegraphics[width=0.85\linewidth]{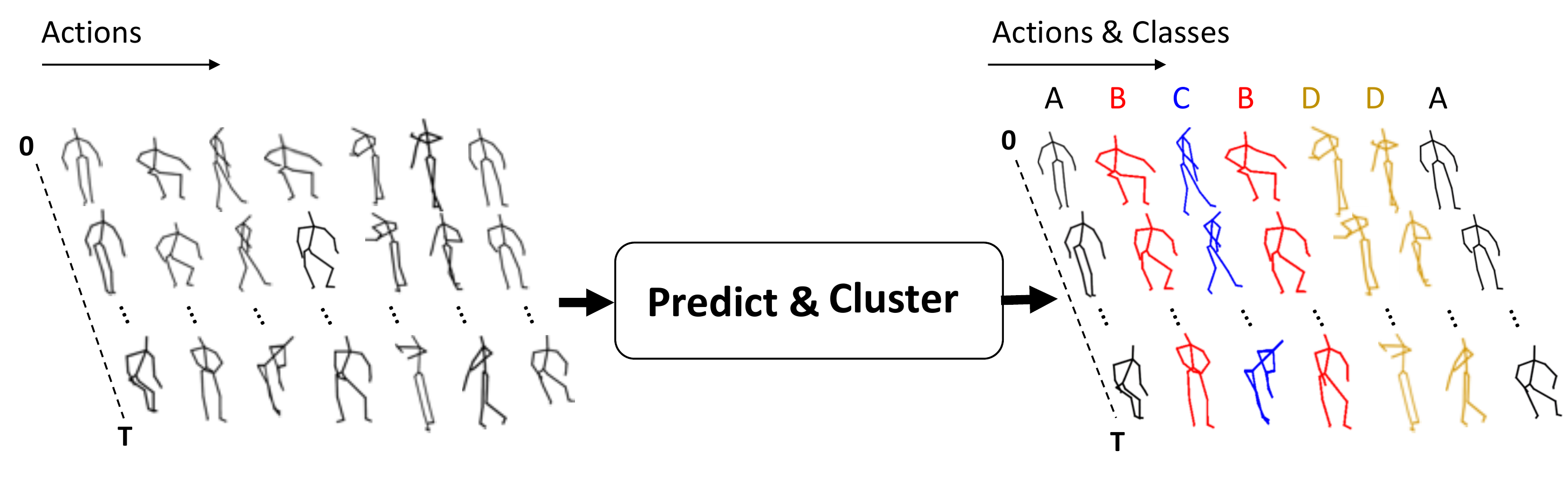}
    \captionof{figure}{PREDICT \& CLUSTER: Unsupervised action recognition from body keypoints. Please see examples in the supplementary video.}\label{fig:teaser}
\end{center}
}]
% \thispagestyle{empty}
%%%%%%%%% ABSTRACT
\begin{abstract}
We propose a novel system for unsupervised skeleton-based action recognition. Given inputs of body keypoints sequences obtained during various movements, our system associates the sequences with actions. Our system is based on an encoder-decoder recurrent neural network, where the encoder learns a separable feature representation within its hidden states formed by training the model to perform prediction task. We show that according to such unsupervised training the decoder and the encoder self-organize their hidden states into a feature space which clusters similar movements into the same cluster and distinct movements into distant clusters. Current state-of-the-art methods for action recognition are strongly supervised, i.e., rely on providing labels for training. Unsupervised methods have been proposed, however, they require camera and depth inputs (RGB+D) at each time step. In contrast, our system is fully \textit{unsupervised}, does not require labels of actions at any stage, and can operate with \textit{body keypoints} input only. Furthermore, the method can perform on various dimensions of body keypoints (2D or 3D) and include additional cues describing movements. We evaluate our system on three extensive action recognition benchmarks with different number of actions and examples. Our results outperform prior unsupervised skeleton-based methods, unsupervised RGB+D based methods on cross-view tests and while being unsupervised have similar performance to supervised skeleton-based action recognition.
\end{abstract}

%%%%%%%%% BODY TEXT
\section{Introduction}
Robust action recognition, especially human action recognition, is a fundamental capability in ubiquitous computer vision and artificial intelligence systems. While recent methods have shown remarkable success rates in recognizing basic actions in videos, current methods rely on strong supervision with a large number of training examples accompanied with action labels. Collection and annotation of large scale datasets is implausible for various types of actions and applications. Furthermore, annotation is a challenging problem by itself, since it is often up to the interpretation of the annotator to assign a meaningful label for a given sequence. This is particularly the case in situations where it is unclear what is the ground truth label, e.g., annotation of animal movements. Indeed, annotations challenges are common in different contextual information on movement, such as video (RGB), depth (+D) and keypoints tracked over time. Compared to RGB+D data, keypoints include much less information and can be challenging to work with. However, on the other hand, focusing keypoints can often isolate the actions from other information and provide more robust unique features for actions. 

For human action recognition, time-series of body joints (skeleton) tracked over time are indeed known as effective descriptors of actions. Here we focus on $3D$ skeleton time sequences and propose an unsupervised system to learn features and assign actions to classes according to them. We call our system PREDICT \& CLUSTER (P\&C) since it is based on training an encoder-decoder type network to both predict and cluster skeleton sequences such that the network learns an effective hidden feature representation of actions. Indeed, an intuitive replacement of a classification supervised task by a non-classification unsupervised task is to attempt to continue (predict) or reproduce (re-generate) the given sequence such that it leads the hidden states to capture key features of the actions. In the encoder-decoder architecture, the prediction task is typically implemented as follows: given an action sequence as the encoder input, the decoder continues or generates the encoder input sequence. Since inputs are sequences, both the decoder and the encoder are recurrent neural networks (RNN) containing cells with hidden variables for each time sample in a sequence. The final hidden state of the encoder is typically being utilized to represent the action feature. While the encoder contains the final action feature, since the gradient during training flows back from the decoder to the encoder, it turns out that the decoder training strategies significantly determine the effectiveness of the representation. Specifically, there are two types of decoder training strategies proposed for such prediction/re-generation task~\cite{srivastava2015unsupervised}. The first strategy is a conditional strategy, where the output of the previous time-step of the decoder is used as input to the current time-step. With such strategy the output of the decoder is expected to be continuous. In contrast, the unconditional strategy assigns a zero input into each time-step of the decoder. Previous work showed that unconditional training of the decoder is expected to have better prediction performance since it effectively weakens the decoder and thus forces the encoder to learn a more informative representation. 

In our system, we extend such strategies to enhance the encoder representation. This results in enhanced clustering and organization of actions in the feature space. In particular, we propose two decoder training strategies, Fixed Weights and Fixed States to further penalize the decoder. The implementation of these strategies guides the encoder to further learn the feature representation of the sequences that it processes. In fact, in both strategies, the decoder is a `weak decoder', i.e., the decoder is effectively not being optimized and it serves the role of propagating the gradient to the encoder to further optimize its final state. Combining these two strategies together, we find that the network can learn a robust representation and our results show that this strategy can achieve significantly enhanced performance than unsupervised approaches trained without them. We demonstrate the effectiveness and the generality of our proposed methodology by evaluating our system on three extensive skeleton-based and RGB+D action recognition datasets. Specifically, we show that our P\&C unsupervised system achieves high accuracy performance and outperforms prior methods.
%-------------------------------------------------------------------------
\begin{figure*}[!t]
    \begin{center}
        \includegraphics[width=0.9\textwidth]{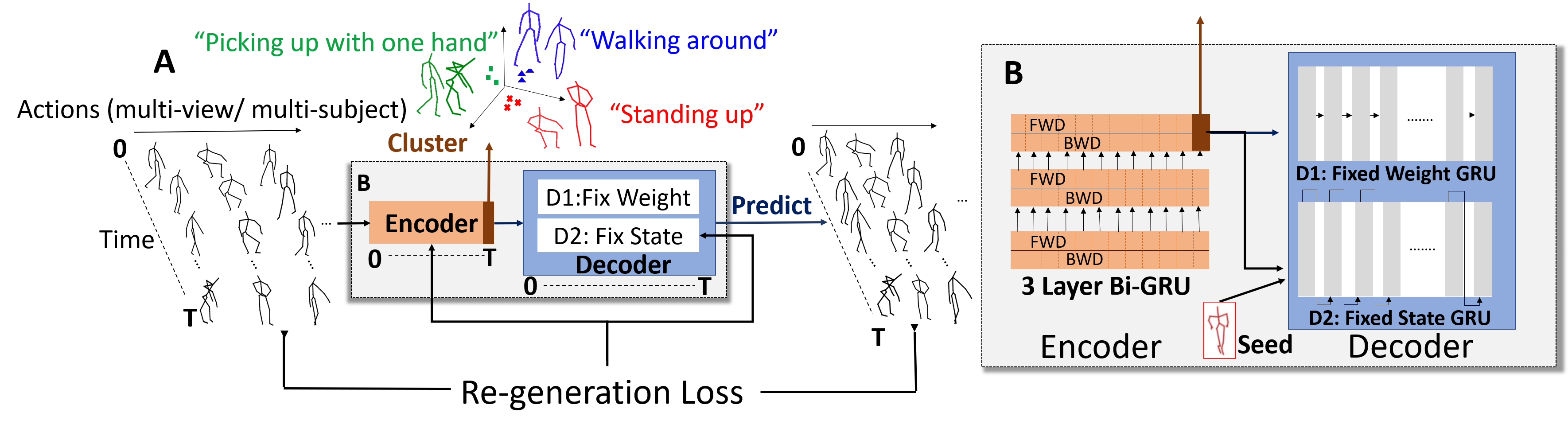}
        \caption{PREDICT \& CLUSTER system summary. \textbf{A:} System overview. \textbf{B:} Encoder-Decoder architecture.}
        \label{fig:sysoverview}
    \end{center}
\end{figure*}
\section{Related Work}
The objective of action recognition is to assign a class label to a sequence of frames with context information on the action performed, Fig.~\ref{fig:teaser}. Numerous approaches have been introduced particularly for human movement action recognition. Such approaches use video frames (RGB), and/or depth (RGB+D) and/or skeleton data, i.e. tracking of body joints (keypoints). Performing exclusive skeleton-based action recognition is especially advantageous since requires much less data, which is relatively easy to acquire and therefore has the potential to be performed in real-time.  Furthermore, in contrast to videos and depth, including various contexts such as background, skeleton data can be used to understand the exclusive features of the actions. Indeed, in recent years, various supervised and unsupervised approaches have been introduced for human skeleton-based action recognition. Most of skeleton-based approaches have been supervised approaches where an annotated set of actions and labels should be provided for training them. In an unsupervised setup, the problem of action recognition is much more challenging. Only a few unsupervised skeleton-based approaches have been proposed and several unsupervised approaches have been developed to use more information such as video frames and depth, i.e. unsupervised RGB+D. We review these prior approaches below and compare our results with them.

For \textit{supervised} skeleton-based action recognition, prior to deep learning methods, classical approaches were proposed to map the actions from Lie group to its Lie algebra and to perform classification using a combination of dynamic time warping, Fourier temporal pyramid representation and linear SVM (e.g. LARP~\cite{vemulapalli2014human}). Deep learning approaches have been developed to classify skeleton data as well, in particular, RNN based models that are designed to work with sequences. For example, Du et al.~\cite{du2015hierarchical} used hierarchical RNN (HBRNN-L) for action classification and Shahroudy et al.~\cite{shahroudy2016ntu} proposed part-aware LSTM (P-LSTM) as a baseline for the large scale skeleton action recognition NTU RGB+D dataset. Since skeleton data is noisy, largely due to variance in camera views, Zhang et al.~\cite{zhang2019view} proposed a view-adaptive RNN (VA-RNN) which learns a transformation from original skeleton data to a general pose. CNN based approaches have been also proposed for supervised skeleton based recognition by constructing a representation of body joints that can be processed by CNN. In particular, Du et al.~\cite{du2015skeleton} represented a skeleton sequence as a matrix by concatenating the joint coordinates in each instant and arranging those vector representations in a chronological order and transforming the matrix into an image on which CNN is trained for classification. In addition, Liu et al.\cite{liu2017enhanced} used an enhanced skeleton visualization method in conjunction with CNN classification for view invariant human action recognition. Recently, graph convolution networks (GNN) gained popularity in skeleton-based action recognition approaches. Yan et al.~\cite{yan2018spatial} introduced Spatial Temporal Graph Convolutional Networks (ST-GCN), which were shown to be capable to learn both the spatial and temporal patterns from skeleton data. A recent extension of such an approach by Shi et al.~\cite{shi2019skeleton},\cite{shi2019two} showed that directed GNN can be used to encode the skeleton representation and also showed that two-stream GNN can learn the graph in an adaptive manner.

While recent supervised approaches show robust performance on action recognition, the \textit{unsupervised} setup is advantageous since it does not require labeling of sequences and may not need re-training when additional actions, not included in the training set, are introduced. Unsupervised methods typically aim to obtain an effective feature representations by predicting future frames of input action sequences or by re-generating the sequences.
Unsupervised approaches were mostly proposed for videos of actions or videos with additional information such as depth or optical flow. Specifically, Srivastava et al.~\cite{srivastava2015unsupervised} proposed a recurrent-based sequence to sequence (Seq2Seq) model as an autoencoder to learn the representation of a video. Such an approach is at the core of our method for body joints input data. However, as we show, the approach will not be able to achieve efficient performance without particular training strategies that we develop to weaken the decoder and strengthen the encoder. Subsequently, Luo et al.~\cite{luo2017unsupervised} developed a convolutional LSTM to use depth and optical flow information such that the network encodes depth input and uses the decoder to predict the optical flow of future frames. Furthermore, Li et al.~\cite{li2018unsupervised} proposed to employ a generative adversarial network (GAN) with a camera-view discriminator to assist the encoder in learning better representations.

As in unsupervised RGB+D approaches, skeleton-based approaches utilize the task of human motion prediction as the underlying task to learn action feature representation. For such a task, RNN-based Seq2Seq models~\cite{martinez2017human} were shown to achieve improved accuracy in comparison to non-Seq2Seq based RNN models such as ERD~\cite{fragkiadaki2015recurrent} and S-RNN~\cite{jain2016structural}. Recently, networks incorporating GANs have achieved improved performance on this task by utilizing the predictor network being RNN Seq2Seq and the discriminator~\cite{gui2018adversarial}.

\textit{Unsupervised} approaches for \textit{skeleton-based action recognition} are scarce since obtaining effective feature representations from coordinate positions of body joints is challenging. In particular, based on successful human motion prediction network configurations, Zheng et al. \cite{zheng2018unsupervised} (LongT GAN)  proposed a GAN encoder-decoder such that the decoder attempts to re-generate the input sequence and the discriminator is used to discriminate whether the re-generation is accurate. The feature representation used for action recognition is taken from the final state of the encoder hidden representation. During training, the masked ground truth input is provided to the decoder. The method was tested on motion-capture databases, e.g., CMU Mocap, HDM05\cite{cg-2007-2} and Berkeley MHAD\cite{ofli2013berkeley}. Such datasets were captured by physical sensors (markers) and thus are much cleaner than marker-less data collected by depth cameras and do not test for  multi-view variance which significantly affects action recognition performance. Our baseline network architecture is similar to the structure in Zheng et al.~\cite{zheng2018unsupervised} since we use an encoder and decoder and we also use the final state of the encoder as a features representation of the action sequences. However, as we show, it is required to develop extended training strategies for the system to be applicable to larger scale multi-view and multi-subject datasets. Specifically, instead of using the masked ground truth as an input into the decoder, we propose methods to improve learning of the encoder and to weaken the decoder.
%\\
%\subsection{Human motion prediction}
%------------------------------------------------------------------------
\begin{figure}[!t]
    \includegraphics[width=\linewidth]{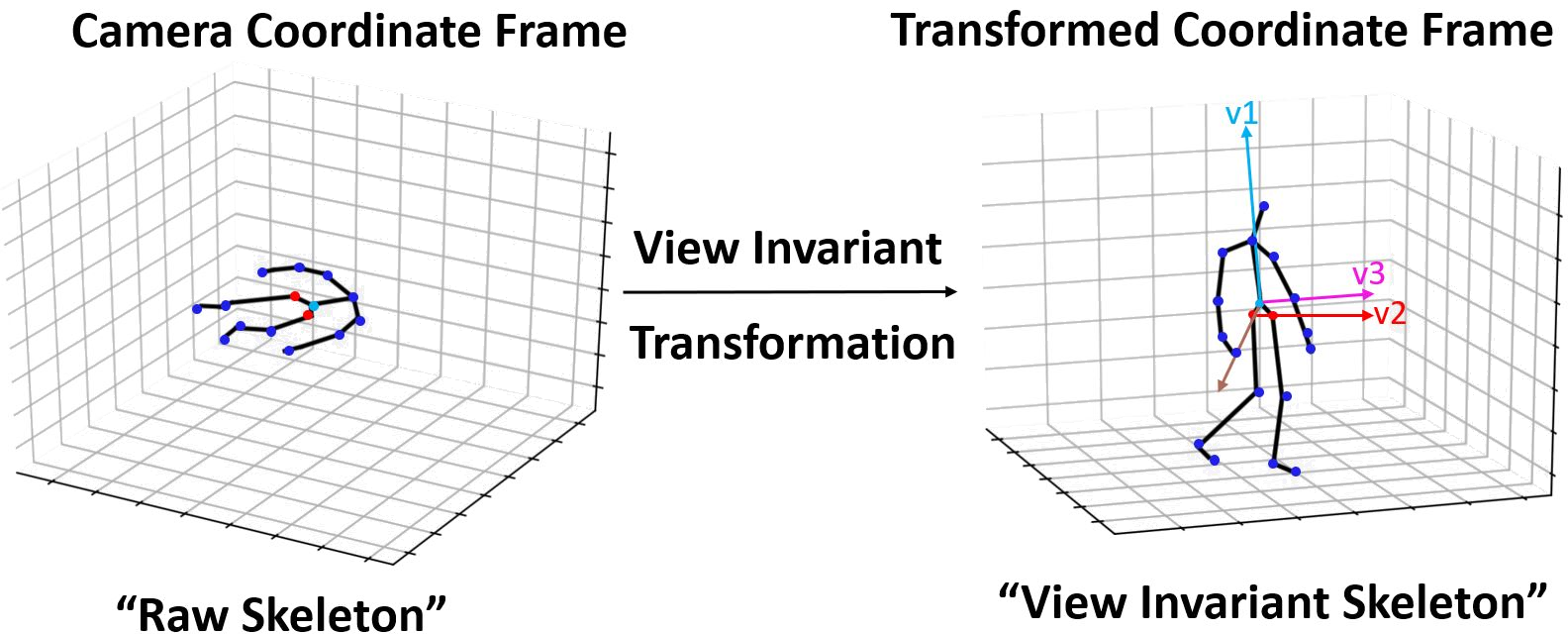}
    \caption{Pre-processing of body keypoints sequences according to view-invariant transformation.}
    \label{fig:preprocess}
\end{figure}
\section{Methods}\label{sec:methods}
\textbf{Pre-processing of body keypoints}: Body keypoints data is a sequence $X^V_T$ of $T$ frames captured from a particular view, where each frame represents  $N=3$D coordinates of $J$ joint keypoints
\begin{align*}
    X_T^V = \{ x_1, x_2, \dots, x_T\}, x_t\in \mathbf{R}^{J\times N}.
\end{align*}
Action sequences are captured from different views by depth camera such as Microsoft Kinect and $3$D human joint positions are extracted from a single depth image by a real-time human skeleton tracking framework\cite{shotton2011real}. We align the action sequences by implementing a view-invariant transformation which transforms the keypoints coordinates from original coordinate system into a view-invariant coordinate system $X^V \to X$. The transformed skeleton joint coordinates are given by
\begin{align*}
    x^j_t = R^{-1}(x^j_t - d_R), \forall j \in J, \forall t\in T,
\end{align*}
where $x^j_t \in \mathbf{R}^{3\times 1}$ are the coordinates of the $j$-th joint of the $t$-th frame, $R$ is the rotation matrix and $d_R$ is the origin of rotation. These are computed according to
\begin{align*}
    R &= \left[ \frac{v_1}{\Vert{v_1}\Vert} \left| \frac{\hat{v}_2}{\Vert{\hat{v}_2}\Vert} \right| \frac{v_1 \times \hat{v}_2}{\Vert{v_1 \times \hat{v}_2}\Vert} \right],d_R = x^{\text{root}}_{t=0},
\end{align*}
where $v_1 = x^{\text{spine}}_{t=0} - x^{\text{root}}_{t=0}$ is the vector perpendicular to the ground, $v_2=x^{\text{hip left}}_{t=0} - x^{\text{hip right}}_{t=0}$ is the difference vector between left and right hips joints in the initial frame of each sequence and $\hat{v}_2 =\frac{v_2-\text{Proj}_{v_1}(v_2)}{\Vert v_2-\text{Proj}_{v_1}(v_2)\Vert}$. $\text{Proj}_{v_1}(v_2)$ and $v_1 \times \hat{v}_2$ denotes the vector projection of $v_2$ onto $v_1$ and the cross product of $v_1$ and $\hat{v}_2$, respectively. $x^{\text{Root}}_{t=0}$ is the coordinate of the root joint in the initial frame \cite{lee2017ensemble} (see Fig.~\ref{fig:preprocess}). Since actions can be of different lengths we down-sample each action sequence to be at most a fixed length $T_{max}$ and pad with zeros if the sequence length is smaller than that. \\
\textbf{Self-organization of hidden states clustering:}
\begin{figure}[!t]
    \centering
    \includegraphics[width=\linewidth]{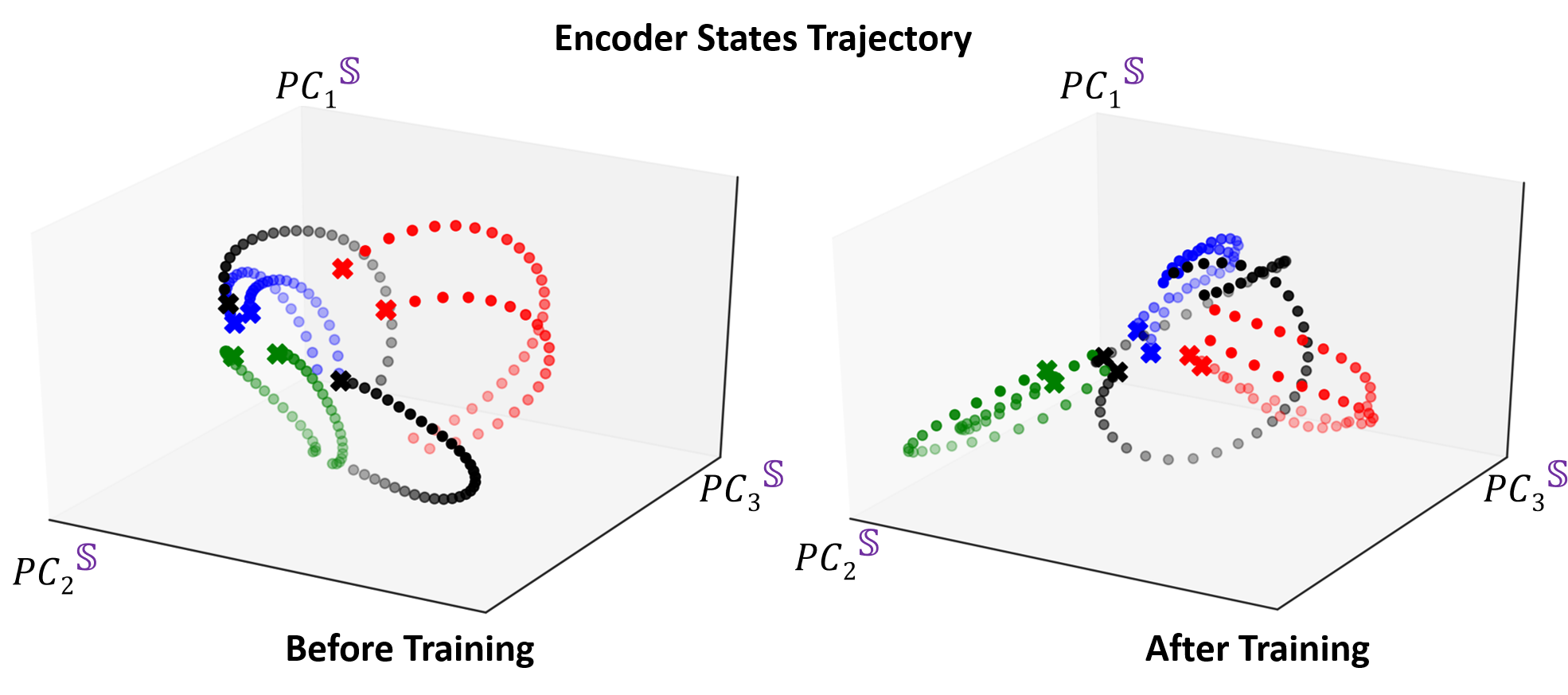}
    \caption{Encoder states trajectories visualized by projection to 3 PCA space. Each color represents one type of action(blue: donning, red: sit down,green:carry, black: stand up). The cross symbol denotes the final state. Left: before training; Right: after training.}
    \label{fig:encoderstates}
\end{figure}
A key property that we utilize in our system is the recent observation that propagation of input sequences through RNN self-organizes them into clusters within the hidden states of the network, i.e., clusters represent features in an embedding of the hidden states\cite{farrell2019recurrent}. Such strategy is a promising unsupervised method for multi-dimensional sequence clustering such as body keypoints sequences~\cite{suclustering}. As we show, self-organization is inherent to any RNN architecture and even holds for random RNN which are initialized with random weights and kept fixed, i.e., no training is performed. Indeed, when we input sequences of body keypoints of different actions into random RNN, the features in the hidden state space turn out to be effective filters. While such strategy is promising, the recognition accuracy outcome appears to be non-optimal (Table \ref{tab:Perf_Table} P\&C Rand). We therefore implement an encoder-decoder system, which we call PREDICT \& CLUSTER (P\&C), where the encoder propagates input sequences and passes the last hidden state to the decoder. The decoder is used to regenerate the encoder input sequences. Furthermore, we utilize the random network setup (which does not require training) to choose the optimal hyper-parameters for the network to be trained. We describe the components of P\&C below.\\
\textbf{Motion prediction}: At the core of our unsupervised method is an encoder-decoder RNN (Seq2Seq). Such network models were shown to be effective at prediction of future evolution of multi-dimensional time-series of features including skeleton temporal data of various actions~\cite{martinez2017human},\cite{gui2018adversarial}. In these applications the typical flow in the network is uni-directional. The \textit{encoder} processes an initial sequence of activity and passes the last state to the \textit{decoder} which in turn, based on this state generates the evolution forward. We extend such network structure for our method (see system overview in Fig.~\ref{fig:sysoverview}).

We propose a \textit{bi-directional} flow such that the network can capture better long-term dependencies in the action sequences. Specifically, the encoder is a multi-layered bi-directional Gated Recurrent Unit (GRU) which input is a whole sequence of body keypoints corresponding to an action. We denote the forward and backward directions hidden states of the last layer of encoder at time $t$ as $ \overrightarrow{E_t}$ and $\overleftarrow{E_t}$ respectively, and the final state of the encoder as their concatenation $E_T = \{\overrightarrow{E_T} , \overleftarrow{E_T} \}$. The decoder is a uni-directional GRU with hidden states at time $t$ denoted as $D_t$. The final state of the encoder is fed into the decoder as its initial state, i.e., $D_0 = E_T$. In such a setup, the decoder generates a sequence based on $E_T$ initialization. In a typical prediction task, the generated sequence will be compared with forward evolution of the same sequence (prediction loss). In our system, since our goal is to perform action recognition, the decoder is required to re-generate the whole input sequence (re-generation loss). Specifically, for the decoder outputs $\hat{X} = \{\hat{x}_1, \hat{x}_2, \dots, \hat{x}_T\}$ the re-generation loss function is the error between $X$ and $\hat{X}$. In particular, we use mean square error (MSE) $L=\frac{1}{T}\sum^{T}_{t=1}(x_t-\hat{x}_t)^2$ or mean absolute error (MAE) $L=\frac{1}{T}\sum^{T}_{t=1} \left|x_t-\hat{x}_t\right|$ as plausible losses.\\
\textbf{Hyper-parameter search}: As in any deep learning system, hyper-parameters significantly impact network performance and require tuning for optimal regime. We utilize the self-organization feature of random initialized RNN to propagate the sequences through the network and use network performance prior to training as an optimization for hyper-parameter tuning. Specifically, we evaluate the capacity of the encoder by propagating the skeleton sequence through the encoder and evaluate the performance of recognition on the final encoder state. We observe that this efficient hyper-parameter search significantly reduces total training time when an optimal network amenable for training is being selected. \\
\textbf{Training}: With optimal hyper-parameter encoder being set, training is performed on the outputs of the decoder to predict (re-generate) the encoder's input action sequence. Training for prediction is typically performed according to one of the two approaches: (i) \textit{unconditional} training in which zeros are being fed into the decoder at each time step or (ii) \textit{conditional} in which an initial input is fed into the first time-step of the decoder and subsequent time-steps use the predicted output of the previous time step as their input \cite{srivastava2015unsupervised}. Based on these training strategies, we propose two decoder configurations \textit{(i) Fixed Weights decoder (FW)} or \textit{(ii) Fixed States decoder (FS)} to weaken the decoder, i.e. to force it to perform the re-generation based upon the information provided by the hidden representation of the encoder and thus improve the encoder's clustering performance, see Fig.\ref{fig:encoderstates}.
\\
\textit{1.Fixed Weights decoder} (\textbf{FW}): The input into the decoder is unconditional in this configuration. The decoder is not expected to learn a useful information for prediction and it exclusively relies on the state passed by the encoder. The weights of the decoder can thereby be assigned as random and the decoder is used as a recurrent propagator of the sequences. In training for the re-generation loss such configuration is expected to \textit{force the encoder to learn the latent features} and represent them with the final state passed to the decoder. This intuitive method turns out to be computationally efficient since only the encoder is being trained and our results indicate favorable performance in conjunction with KNN action classification.
\\\textit{2.Fixed States decoder} (\textbf{FS}): The external input into the decoder is conditional in this configuration ( external input into each time-step is the output of the previous time-step), however the internal input, typically the hidden state from previous step, is replaced by the final state of the encoder $E_T$. Namely, in RNN cell
\begin{align*}
        &h_t = \sigma (W_x x_t + W_h g_{t}+b_h), g_{t}=h_{t-1} \to E_T, \\ 
        &y_t = \sigma (W_y h_t+b_y),\\
        &x_{t+1} = y_t,
\end{align*}
with $x_t$ the external input, $y_t$ the output and $h_t$ the hidden state at time-step $t$, $h_{t-1}$ terms are replaced by $E_T$. In addition, we also add residual connection between external input and output, which has been shown useful in human motion prediction as well~\cite{martinez2017human}. The final output and next input will be $\hat{y}_t = y_t + x_t$ and $\hat{x}_{t+1}=\hat{y}_t$, respectively. The configuration forces the network to rely on $E_T$, instead of the hidden state at previous time-step and eliminates vanishing of the gradient since during back-propagation at each time-step there is a defined gradient back to the final encoder state.
\\
\begin{figure}[!t]
    \includegraphics[width=0.9\linewidth]{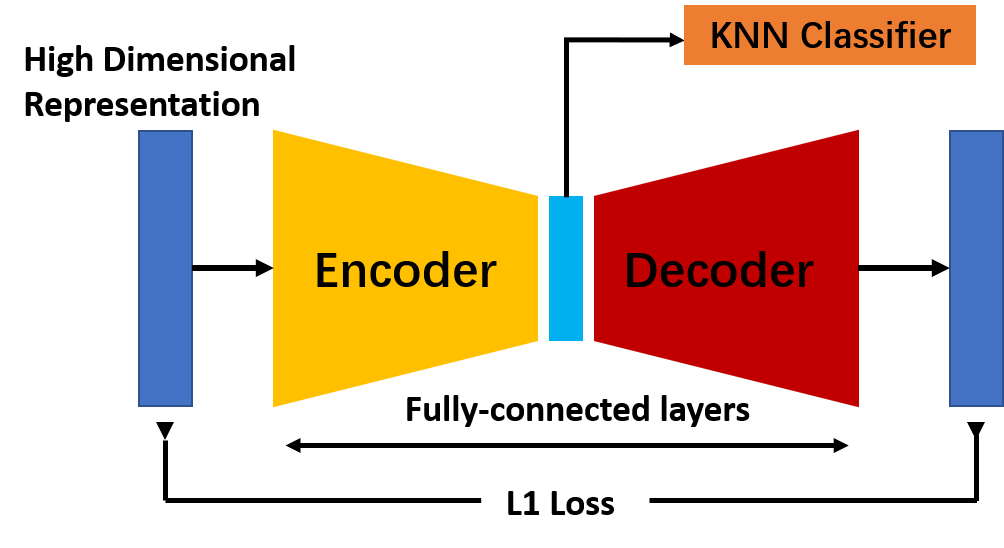}
    \caption{Feature-level autoencoder and KNN Classifier}
    \label{fig:classify}
\end{figure}
\begin{figure*}[t]
    \centering
    \includegraphics[width=0.95\linewidth]{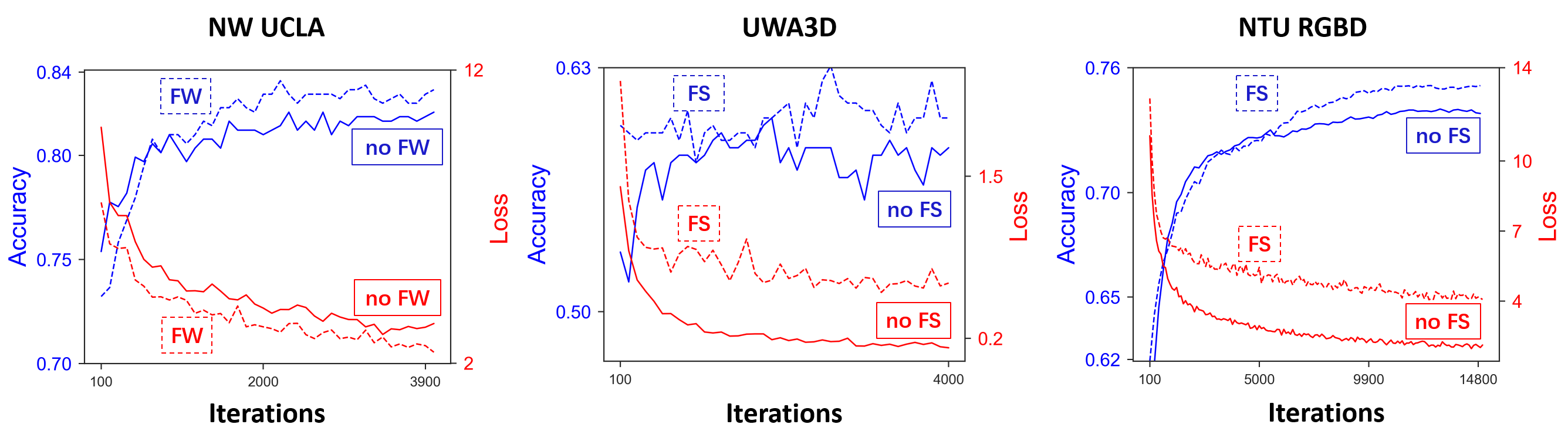}
    \caption{Training curves (accuracy:blue; loss:red) for three of the datasets, left to right: NW-UCLA (FW v.s. no FW), UWA3D(FS v.s. no FS), NTU-RGB+D Cross-View(FS v.s. no FS).}
    \label{fig:all_acc}
\end{figure*}
\textbf{Feature level auto-encoder}: After training the prediction network we extract the final encoder state $E_T$ as the feature vector associated with each action sequence. Since the feature vector is high-dimensional, we use a feature-level auto-encoder that learns the core low dimensional components of the high-dimensional feature so it can be utilized for classification~(Fig.~\ref{fig:classify}). Specifically, we implement the auto-encoder, denoted as $f$ to be of an encoder-decoder architecture with parameters $\theta$ such that 
\begin{equation*}
\hat{E_T} =  f_{\theta}(E_T)\approx E_T.    
\end{equation*} 
The encoder and the decoder are multi-layer FC networks with non-linear $\tanh$ activation and we implement the following loss $l_{aec}=\left| E_T-\hat{E_T} \right|$.\\

\textbf{K-nearest neighbors classifier}: For evaluation of our method on action recognition task we use a K-nearest neighbors (KNN) classifier on the middle layer of the auto-encoder feature vector. Specifically, we apply the KNN classifier (with $k=1$) on the features of the trained network on all sequences in the training set to assign classes. We then use cosine similarity as the distance metric to perform recognition, i.e., place each tested sequence in a class. Notably, KNN classifier does not require to learn extra weights to action placement.
%------------------------------------------------------------------------
\begin{table*}[]
\resizebox{0.29\textwidth}{!}{%
\begin{tabular}[t]{|l|c|}
\hline
 \multirow{2}{*}{Method} & \textbf{NW-UCLA}\\
  & (\%) \\ \hline
\multicolumn{2}{c}{\vspace{-0.25cm}}\\\hline
\multicolumn{2}{|c|}{\textbf{\textcolor{blue}{Supervised Skeleton}}}\\\hline
HOPC\cite{rahmani2014hopc}  & 74.2        \\ 
Actionlet Ens \cite{wang2013learning} & 76.0        \\  
HBRNN-L\cite{du2015hierarchical}  & 78.5        \\ 
VA-RNN-Aug\cite{zhang2019view}    & \textbf{\textcolor{blue}{90.7}} \\ 
AGC-LSTM\cite{si2019attention}  & \textbf{\textcolor{blue}{93.3}}  \\ \hline 
\multicolumn{2}{c}{\vspace{-0.25cm}}                \\\hline
\multicolumn{2}{|c|}{\textbf{\textcolor{purple}{Unsupervised RGB+D}} }\\\hline
Luo et al.\cite{luo2017unsupervised}   & \textcolor{purple}{\textbf{50.7}}       \\ 
Li et al.\cite{li2018unsupervised}     & \textcolor{purple}{\textbf{62.5}}        \\ \hline
\multicolumn{2}{c}{\vspace{-0.25cm}}                 \\\hline
\multicolumn{2}{|c|}{\textbf{\textcolor{red}{Unsupervised Skeleton}} }\\\hline
P\&C Rand \textbf{(Our)}      & 72.0        \\  
LongT GAN~\cite{zheng2018unsupervised}      & 74.3        \\  
% Fixed state                             & 82.3       \\ 
% w.o. fixed dec. state                        & 81.8       \\ 
% & Fixed dec. weight                            & 83.6       \\ 
% & w.o. fixed dec. weight                       & 82.9       \\ 
P\&C FS-AEC \textbf{(Our)}                        & \textbf{\textcolor{red}{83.8}}           \\ 
P\&C FW-AEC \textbf{(Our)}                      & \textbf{\textcolor{red}{84.9}}            \\ \hline
\end{tabular}%
}
\hfill
% \begin{table}[]
\resizebox{0.33\textwidth}{!}{%
\begin{tabular}[t]{|l|c|c|}
\hline
\multirow{2}{*}{Method} & \multicolumn{2}{c|}{\textbf{UWA3D}} \\\cline{2-3} 
\multirow{1}{*}{}  & V3 (\%)   & V4 (\%) \\ \hline \multicolumn{3}{c}{\vspace{-0.25cm}}\\\hline
 \multicolumn{3}{|c|}{\textbf{\textcolor{blue}{Supervised Skeleton}}}\\\hline
HOJ3D\cite{xia2012view} & 15.3 & 28.2                \\
2-layer P-LSTM\cite{wang2019comparative} & 27.6 &24.3\\
IndRNN (6 layers)\cite{wang2019comparative} &30.7 &47.2\\
IndRNN (4 layers)\cite{wang2019comparative} &34.3 &54.8\\
ST-GCN\cite{wang2019comparative}  & 36.4   & 26.2                \\ 
Actionlet Ens\cite{wang2013learning}      & 45.0   & 40.4   \\  
LARP\cite{vemulapalli2014human}   & 49.4 & 42.8  \\
HOPC\cite{rahmani2014hopc}   & \textbf{\textcolor{blue}{52.7}} & \textbf{\textcolor{blue}{51.8}}  \\
VA-RNN-Aug\cite{zhang2019view}  & \textbf{\textcolor{blue}{70.9}}   & \textbf{\textcolor{blue}{73.2}}   \\ \hline
 \multicolumn{3}{c}{\vspace{-0.25cm}}     \\\hline
 \multicolumn{3}{|c|}{\textbf{\textcolor{red}{Unsupervised Skeleton}} }\\\hline
P\&C Rand \textbf{(Our)}      & 48.5                    & 51.5                   \\  
LongT GAN~\cite{zheng2018unsupervised}      & 53.4     & 59.9  \\  
% Fixed state             & 58.7                 & 63.0                \\
% w.o. fixed dec. state        & 54.6                & 60.3           \\
% Fixed dec. weight            & 58.7                 & 62.3          \\ w.o. fixed dec. weight       & 57.1                & 60.3            \\
P\&C FS-AEC \textbf{(Our)} & \textbf{\textcolor{red}{59.5}} & \textbf{\textcolor{red}{63.1}}                   \\ 
P\&C FW-AEC \textbf{(Our)} & \textbf{\textcolor{red}{59.9}}                     & \textbf{\textcolor{red}{63.1}}                  \\ \hline
\end{tabular}%
}
% \caption{Performance on UWA3D dataset (Test set of 30 Classes)}
% \label{UWA3D_table}
%}
% \end{table*}
\hfill
% \begin{table*}[]
% \centering
\resizebox{0.35\textwidth}{!}{%
\begin{tabular}[t]{|l|c|c|}
\hline
\multirow{2}{*}{Method} & \multicolumn{2}{c|}{\textbf{NTU RGB-D 60}} \\\cline{2-3} 
\multirow{1}{*}{}  &  C-View  & C-Subject  \\ 
\multirow{1}{*}{}  &  (\%)  & (\%)  \\ \hline
 \multicolumn{3}{c}{\vspace{-0.25cm}}\\\hline
 \multicolumn{3}{|c|}{\textbf{\textcolor{blue}{Supervised Skeleton}}}\\\hline
HOPC\cite{rahmani2014hopc} & 52.8 & 50.1 \\ 
HBRNN\cite{du2015hierarchical}   & 64.0 & 59.1 \\ 
2L P-LSTM\cite{shahroudy2016ntu} & 70.3 & 62.9 \\ 
ST-LSTM\cite{liu2016spatio}  & \textbf{\textcolor{blue}{77.7}} & \textbf{\textcolor{blue}{69.2}}    \\
VA-RNN-Aug\cite{zhang2019view}   & \textbf{\textcolor{blue}{87.6}}& \textbf{\textcolor{blue}{79.4}}    \\ \hline
 \multicolumn{3}{c}{\vspace{-0.25cm}}\\\hline
 \multicolumn{3}{|c|}{\textbf{\textcolor{purple}{Unsupervised RGB+D}}}\\\hline
Shuffle \& learn\cite{misra2016shuffle}        & 40.9  & 46.2 \\  
Luo et al.\cite{luo2017unsupervised}              & \textbf{\textcolor{purple}{53.2}} &\textbf{\textcolor{purple}{61.4}}  \\ 
Li et al.\cite{li2018unsupervised} & \textbf{\textcolor{purple}{63.9}} & \textbf{\textcolor{purple}{68.1}}\\ \hline
 \multicolumn{3}{c}{\vspace{-0.25cm}}                   \\\hline
 \multicolumn{3}{|c|}{\textbf{\textcolor{red}{Unsupervised Skeleton}} }\\\hline
LongT GAN~\cite{zheng2018unsupervised}  & 48.1 & 39.1      \\  
P\&C Rand \textbf{(Our)}    & 56.4 & 39.6                           \\  
% Fixed state             & 49.2                 & 75.3                \\
% w.o. fixed dec. state        & 47.3                & 74.0           \\
% Fixed dec. weight            & 50.3                 & 75.2          \\ w.o. fixed dec. weight       & 50.4                & 74.5            \\
P\&C FS-AEC \textbf{(Our)}  & \textbf{\textcolor{red}{76.3}} & \textbf{\textcolor{red}{50.6}}  \\ 
P\&C FW-AEC \textbf{(Our)} & \textbf{\textcolor{red}{76.1}} & \textbf{\textcolor{red}{50.7}} \\ \hline
\end{tabular}%
}
\caption{Comparison of action recognition performance of our P\&C system with state-of-the-art approaches of \textit{Supervised Skeleton} (blue) and \textit{Unsupervised} \textit{RGB+D} (purple); \textit{Unsupervised Skeleton} (red)) types.}
\label{tab:Perf_Table}
\end{table*}
\section{Experimental Results and Datasets}
\textbf{Implementation details}: To train the network, all body keypoints sequences are pre-processed according to the view-invariant transformation and down-sampled to have at most $50$ frames (Fig.~\ref{fig:preprocess}). The coordinates are also normalized to the range of $[-1,1]$. Using the hyper-parameter search, employing random RNN propagation discussed above, we set the following architecture: \textit{Encoder}: $3$-Layer Bi-GRU with $N=1024$ units in each layer. \textit{Decoder}: 1-Layer Uni-GRU with $N=2048$ units such that it is compatible with the dimensions of the encoder final state $E_T$.All GRUs are initialized with a random uniform distribution. \textit{Feature-level auto-encoder:} $6$ FC Layers with the following dimensions: input feature vector(dim$=2048$) $\rightarrow$ FC($1024$) $\rightarrow$ FC($512$) $\rightarrow$ FC($256$)$\rightarrow$ FC($512$)$\rightarrow$FC($1024$) $\rightarrow$FC($2048$). All FCs use $\tanh$ activation except the last layer which is linear. The middle layer of auto-encoder outputs a vector feature of $256$ elements which is used as the final feature. We use Adam optimizer and learning rate starting from $0.0001$ and $0.95$ decay rate at every $1000$ iterations. The gradients are clipped if the norm is greater than $25$ to avoid gradient explosion. It takes $0.7$sec per training iteration and $0.1$sec to forward propagate with batch size of $64$ on one Nvidia Titan X GPU. Please see additional details of architecture choices in the supplementary material.
\\
\textbf{Datasets}: We use three different data-sets for training, evaluation and comparison of our P\&C system with related approaches. The three data-sets include various number of classes, types of actions, body keypoints captured from different views and on different subjects. In these datasets, the body keypoints are captured by depth cameras and also include additional data, e.g., videos (RGB) and depth (+D). Various types of action recognition approaches have been applied to these datasets, e.g., supervised skeleton approaches and unsupervised RGB+D approaches. We list these types of approaches and their performance on the tests in the datasets in Table~\ref{tab:Perf_Table}. Notably, as far as we know, our work is the \textbf{first fully unsupervised skeleton based approach applied on these extensive action recognition tests}.

The datasets that we have applied our P\&C system to are \textit{(i) NW-UCLA}, \textit{(ii) UWA3D}, and \textit{(iii) NTU RGB+D}. The datasets include $3$D body keypoints of $10, 30, 60$ action classes respectively. We briefly describe them below.
\textbf{North-Western UCLA (NW-UCLA)} dataset~\cite{wang2014cross} is captured by Kinect v$1$ and contains $1494$ videos of $10$ actions. These actions are performed by $10$ subjects repeated $1$ to $6$ times. There are three views of each action and for each subject $20$ joints are being recorded. We follow \cite{liu2017enhanced} and \cite{wang2014cross} to use the first two views (V$1$,V$2$) for training and last views (V$3$) to test cross view action recognition.
\textbf{UWA3D Multiview Activity II (UWA3D)} dataset~\cite{rahmani2014hopc} contains $30$ human actions performed $4$ times by $10$ subjects. $15$ joints are being recorded and each action is observed from four views: frontal, left and right sides, and top. The dataset is challenging due to many views and the resulting self-occlusions from considering only part of them. In addition, there is a high similarity among actions, e.g., the two actions "drinking” and “phone answering” have many keypoints being near identical and not moving and there are subtle differences in the moving keypoints such as the location of the hand.
\textbf{NTU RGB+D} dataset~\cite{shahroudy2016ntu} is a large scale dataset for $3$D human activity analysis. This dataset consists of $56,880$ video samples, captured from $40$ different human subjects, using Microsoft Kinect v2. NTU RGB+D($60$) contains $60$ action classes. We use the $3$D skeleton data for our experiments such that each time sample contains $25$ joints. We test our P\&C method on both cross-view and cross-subject protocols.

%------------------------------------------------------------------------
\begin{figure*}[t!]
\centering    
\includegraphics[width=0.8\linewidth]{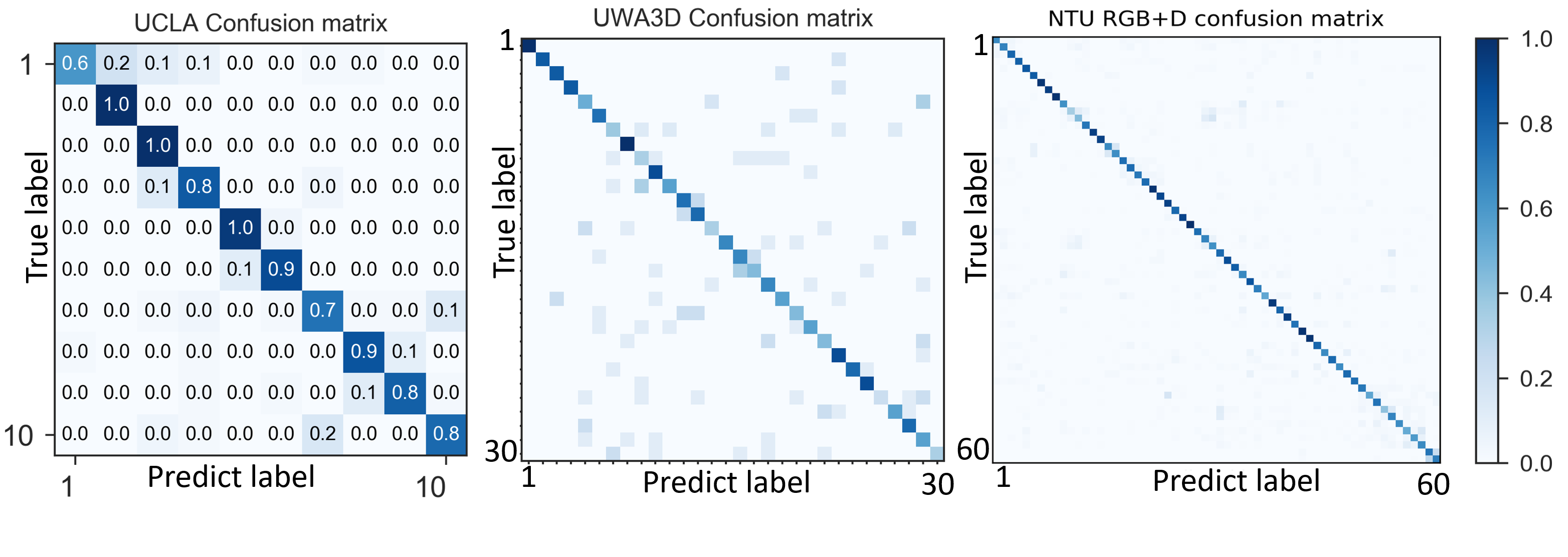}
    \caption{Confusion matrices for testing P\&C performance on the three datasets(from left to right): NW-UCLA(10 actions); UWA3D V4(30 actions); NTU-RGBD Cross-View(60 actions).}
    \label{fig:all_cm}
\end{figure*}
\begin{figure}[!]
\centering    
    \includegraphics[width=0.7\linewidth]{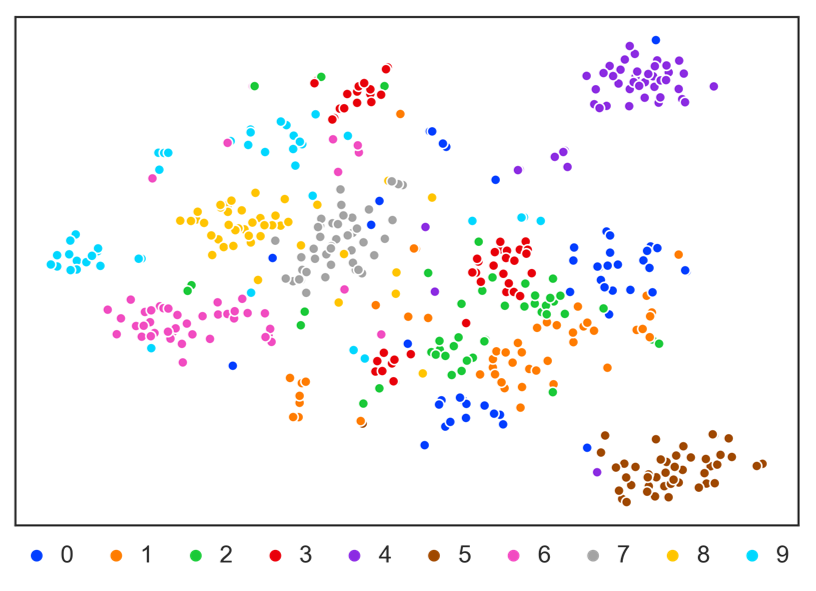}
    \caption{t-SNE visualization of learned features on NW-UCLA dataset.}
    \label{fig:ucla_tsne}
\end{figure}
\section{Evaluation and Comparison}
\textbf{Evaluation}: In all experiments, we use the K-nearest neighbors classifier with $k=1$ to compute the action recognition accuracy and evaluate the performance of our P\&C method. We test different variants of P\&C architectures (combinations of components described in Section ~\ref{sec:methods}) and report a subset of these in the paper: \textit{baseline} random initialized encoder with no training (\textbf{P\&C-Rand}), full system with FS decoder and feature-level auto-encoder (\textbf{P\&C-FS-AEC}) and full system with FW decoder and feature-level auto-encoder (\textbf{P\&C-FW-AEC}). We report the rest of the combinations and their results in the Supplementary material.

Fig.~\ref{fig:all_acc} shows the optimization of the regeneration loss (red) and the resulting accuracy (blue) during training for each dataset. We include plots of additional P\&C configurations in the Supplementary material. The initial accuracy appears to be substantial and this is attributed to the hyper-parameter search being performed on random initialized networks prior to training that we describe in Section \ref{sec:methods}. Indeed, we find that using appropriate initialization, the encoder, without any training, effectively directs similar action sequences to similar final states. 
Training enhances this performance further in both P\&C FW and P\&C FS configurations. Over multiple training iterations both P\&C FW and P\&C FS achieve higher accuracy than no FW and no FS in all datasets. While the convergence of the loss curve indicates improvement on the accuracy, the value of the loss does not necessarily indicate a better accuracy as can be observed from loss and accuracy curves of training on UWA3D and NTU-RGB+D (Fig.~\ref{fig:all_acc} middle, right).
% Notably, typical configurations of encoder-decoder will have no correlation between the prediction loss curve and the accuracy of action recognition, such that the accuracy will remain flat while the loss will converge. 
% The particular network components and architectures that we implement here allow us to correlate between loss and accuracy, i.e. to substantially improve the accuracy during training.

We show the confusion matrices for the three considered datasets in Fig.~\ref{fig:all_cm}. In NW-UCLA (with least classes) we show the elements of the 10x10 matrix. Our method achieves high-accuracy ($>83\%$) on average and there are three actions (pick up with two hands, drop trash, sit down) for which it recognizes with nearly $100\%$ accuracy. We also show in Fig.~\ref{fig:ucla_tsne} a t-SNE visualization of the learned features for NW-UCLA test. Even in this $2$D embedding it is clearly evident that the features for each class are well separated. As more action classes are considered, the recognition becomes a more difficult task and also depends on amount of training data.  For example, while NTU RGB+D has more classes than UWA3D, the recognition accuracy on NTU RGB+D is smoother and results with better performance since it has $40$ times more data than UWA3D. Our results show that our method is compatible with varying data sizes and number of classes.\\
% Similarly, in the case of NTU-RGB+D dataset, using fixed decoder state strategy can gradually improve the accuracy of recognition even the loss is higher. In all cases, we can see the accuracy goes up when the prediction loss goes down but we want to point out getting good prediction does not necessarily mean the encoder learns good features for action recognition because decoder is also possible to support partial functionality for doing prediction. 
\textbf{Comparison}: We compare the performance of our P\&C method with prior related supervised and unsupervised methods applied to (left-to-right): NW-UCLA, UWA3D, NTU RGB+D datasets, see Table~\ref{tab:Perf_Table}. In particular, we compare action recognition accuracy with approaches based on supervised skeleton data (blue), unsupervised RGB+D data (purple) and unsupervised skeleton data (red). For comparison with unsupervised skeleton methods, we implement and reproduce the LongTerm GAN model (LongT GAN) as introduced in \cite{zheng2018unsupervised} and list its performance.

For NW-UCLA, P\&C outperforms previous unsupervised methods (both RGB+D and skeleton based). Our method even outperforms the first three supervised methods listed in Table~\ref{tab:Perf_Table}-left. UWA3D is considered a challenging test for many deep learning approaches since the number of sequences is small, while it includes a large number of classes ($30$). Indeed, action recognition performance of many supervised skeleton approaches is low ($<50\%$). For such datasets, it appears that the unsupervised approach could be more favorable, i.e., even P\&C Rand reaches performance of $\approx 50\%$. LongT GAN achieves slightly higher performance than P\&C Rand, however, not as high as P\&C FS/FW-AEC which perform with $\approx 60\%$. Only a single supervised skeleton method, VA-RNN-Aug, is able to perform better than our unsupervised approach, see Table~\ref{tab:Perf_Table}-middle. On the large scale NTU-RGB+D dataset, our method performs extremely well on the cross-view test. It outperforms prior unsupervised methods (both RGB+D and skeleton based) and on-par with ST-LSTM (second best supervised skeleton method), see Table~\ref{tab:Perf_Table}-right. On the cross-subject test we obtain performance that is higher (including P\&C Rand) than the prior unsupervised skeleton approach, however, our accuracy does not outperform unsupervised RGB+D approaches. We believe that the reason stems from skeleton based approaches not performing well in general on cross-subject tests since additional aspects such as subjects parameters, e.g., skeleton geometry and invariant normalization from subject to subject, need to be taken into account.

In summary, for all three datasets, we used a single architecture and it was able to outperform the prior unsupervised skeleton method, LongT-GAN\cite{zheng2018unsupervised}, most supervised skeleton methods and unsupervised RGB+D methods on cross view tests and some supervised skeleton and unsupervised RGB+D on large scale cross subject test. 

\section{Conclusion}
We presented a novel unsupervised model for human skeleton-based action recognition. Our system reaches enhanced performance compared to prior approaches due to novel training strategies which weaken the decoder and training of the encoder. As a result the network learns more separable representations. Experimental results demonstrate that our unsupervised model can effectively learn distinctive action features on three benchmark datasets and outperform prior unsupervised methods.
% While our model performs well in unsupervised skeleton-based action recognition, two potential future work could be:$1)$to understand how training by back propagation exerts the contraction of hidden representations; $2)$we would like to investigate the improve the performance under unsupervised setting by combining additional information besides only joint coordinates.
{\small
\bibliographystyle{ieee_fullname}
\bibliography{egbib}
}

\end{document}